\documentclass[12pt,runningheads]{llncs}

\usepackage[a4paper, margin=1in]{geometry}
\usepackage{caption}
\usepackage{hyperref}

\usepackage[T1]{fontenc}
\usepackage{graphicx}
\usepackage{amsmath}
\usepackage{amssymb}
\usepackage{mathtools}
\usepackage{booktabs} 
\usepackage{float}
\usepackage{tabularx}
\usepackage{array}

\begin{document}
\title{Peptide2Mol: A Diffusion Model for Generating Small Molecules as Peptide Mimics for Targeted Protein Binding}
%
%
\author{
Xinheng He\inst{1}\# \and
Yijia Zhang\inst{2,3}\# \and
Haowei Lin\inst{4} \and
Xingang Peng\inst{4} \and
Xiangzhe Kong\inst{5} \and
Mingyu Li\inst{6,7} \and
Jianzhu Ma\inst{2,3}\thanks{Corresponding author.}%
}

\authorrunning{X. He et al.}

\institute{
Lingang Laboratory, Shanghai, China
\and
Department of Electronic Engineering, Tsinghua University, Beijing, China
\and
Institute for AI Industry Research (AIR), Tsinghua University, Beijing, China
\and
Institute for Artificial Intelligence, Peking University, Beijing, China
\and
Department of Computer Science and Technology, Tsinghua University, Beijing, China
\and
Department of Pharmaceutical and Artificial-Intelligence Sciences, Shanghai Jiao Tong University School of Medicine, Shanghai, China
\and
Institute of Medical Artificial Intelligence, Shanghai Jiao Tong University, Shanghai, China\\[4pt]
\email{majianzhu@tsinghua.edu.cn}
}
\maketitle              
\begin{abstract}
Structure-based drug design has seen significant advancements with the integration of artificial intelligence (AI), particularly in the generation of hit and lead compounds. However, most AI-driven approaches neglect the importance of endogenous protein interactions with peptides, which may result in suboptimal molecule designs. In this work, we present Peptide2Mol, an E(3)-equivariant graph neural network diffusion model that generates small molecules by referencing both the original peptide binders and their surrounding protein pocket environments. Trained on large datasets and leveraging sophisticated modeling techniques, Peptide2Mol not only achieves state-of-the-art performance in non-autoregressive generative tasks, but also produces molecules with similarity to the original peptide binder. Additionally, the model allows for molecule optimization and peptidomimetic design through a partial diffusion process. Our results highlight Peptide2Mol as an effective deep generative model for generating and optimizing bioactive small molecules from protein binding pockets.

\end{abstract}
\newpage
\section{Introduction}
Small molecules have long been the cornerstone of drug discovery due to their ease of synthesis, cell permeability, oral bioavailability, and cost-effectiveness in manufacturing \cite{wu2024overcoming,vargason2021evolution}. In contrast, peptides, despite their high affinity and specificity for protein targets, often suffer from poor membrane permeability and metabolic instability, which severely limit their therapeutic application \cite{muttenthaler2021trends}. To combine the strength of both modalities, recent strategies aim to transform native peptide or protein binders into small molecules that preserve key binding interactions \cite{brytan2023structural}. This concept has been validated in several landmark cases, such as the conversion of the snake-venom peptide Teprotide into the antihypertensive drug Captopril and the rational design of the HIV protease inhibitor Saquinavir through peptide bond isosteres \cite{odaka2000angiotensin,roberts1990rational}. However, these successes remain isolated, and no systematic or scalable framework exists for general peptide-to-small-molecule conversion. 

With the rapid advancement of artificial intelligence (AI), especially the remarkable success of generative models, drug design has entered a new era \cite{gangwal2024unleashing,hexinheng2025ai,luoshitong2022antigen,lin2025diffbp,huang2024dual,luo20213d}. Early generative models learned structural distributions from known ligands \cite{godinez2022design,bagal2021molgpt}, while recent methods incorporate pocket structures to generate target-specific molecules \cite{lin2025diffbp,jiang2024pocketflow,choi2024pidiff}. This shift is motivated by the recognition that incorporating receptor-specific information is vital for drug design, because only through precise binding to the target protein can a drug exert its therapeutic effect \cite{santos2017comprehensive}. 

Recent advancements in predictive modeling have provided promising approaches for molecular generation. Modern all-atom models have demonstrated the capability to predict small molecule-protein complexes with atomic-level precision \cite{roy2024alphafold3}. Furthermore, work from the Baker group has shown has demonstrated that deep learning can be used to design diverse, high-affinity protein binders \cite{watson2023novo,fox2025code}. However, small molecule generation encompasses a broader chemical space and presents challenges in terms of validation, which can be costly and time-consuming \cite{zeng2022deep}. 

Despite this, most existing models focus solely on small molecule-protein complex data, often overlooking the abundant and biologically significant protein-protein and protein-peptide interactions \cite{greenblatt2024discovery}. This narrow focus on protein-ligand complexes introduces several challenges. This limited scope leads to a lack of diversity in the generated small molecules, as the available protein-ligand complexes often represent similar scaffolds, thereby constraining the exploration of novel molecular designs \cite{zhu2025augmented,zhang2024deep}. Moreover, current models fail to incorporate protein-protein or protein-peptide interaction structural data, despite the growing emphasis on mimicking peptide binders in small molecule design. Consequently, a significant gap exists in the ability to effectively link small molecules and peptides/proteins at the atomic scale for generation. 

To address this gap, we propose Peptide2Mol, the first AI model that learns to translate peptide or protein binding interfaces into small molecules directly in three-dimensional space. Peptide2Mol is formulated as an E(3)-equivariant graph neural network (EGNN) diffusion model, trained on diverse datasets encompassing small-molecule conformation ensembles, protein–ligand complexes, and both experimentally determined and computationally predicted protein–protein interactions. This design enables Peptide2Mol to generate target-aware small molecules that reliably mimic peptide-like binding interactions while maintaining favorable drug-like properties towards the target protein. Such an algorithm does not conflict with existing diffusion based methods \cite{lin2025diffbp,huang2024dual,guan2023targetdiff} and can be combined to generate small molecule binders to mimic peptide behavior. Finally, by analyzing antibody-antigen surface interactions, we identify preferred chemical groups for replacing amino acids, which provides valuable insights into the design of peptide mimicry.

\section{Methods}

\begin{figure}[htb]
\vspace{-20pt}
   \centering
   \includegraphics[width=0.9\linewidth]{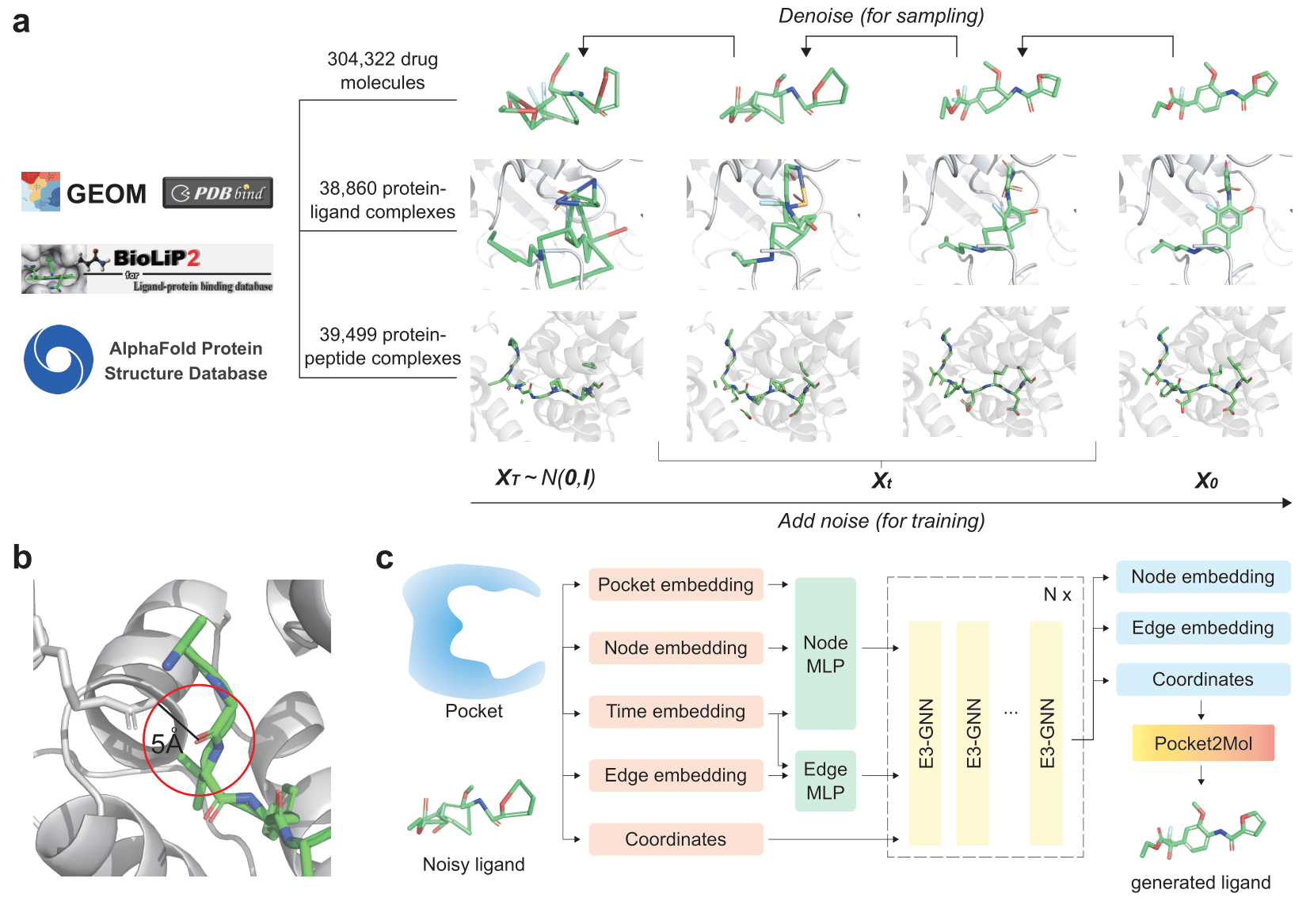}
   \caption{Overview of the Peptide2Mol model. (\textbf{a}) Dataset composition, training, and inference workflow. The model is trained on peptide and small molecule structures, with inference generating candidate ligands for target protein pockets. (\textbf{b}) Schematic representation of edge for non-covalent interactions between ligands and the protein pocket. (\textbf{c}) Model architecture of Peptide2Mol.}
   \label{fig:overview}
   \vspace{-25pt}
\end{figure}

\subsection{Dataset Construction}
To construct our training dataset, we aggregated molecular structures from multiple sources. Small molecules were obtained from the Geometric Ensemble Of Molecules (GEOM) \cite{axelrod2022geom} drug dataset, while protein-ligand and protein-peptide complexes were obtained from the PDBBind \cite{wang2005pdbbind} and BioLip2 \cite{zhang2024biolip2} databases. Protein-peptide interaction were also derived from the monomeric models in the AlphaFold Database \cite{varadi2022alphafold}. In these models, loops that are fully buried and exhibit interactions with other parts are treated as ligands, while the remaining parts are considered receptors. All molecular data were filtered using RDKit \cite{landrum2013rdkit} to ensure successful parsing, yielding a final dataset comprising 304,322 drug-like small molecules, 38,860 protein-ligand complexes, and 39,499 peptide-protein interfaces.

For evaluation, we assembled a test set comprising 10 protein-ligand complexes randomly selected from the CrossDock2020 dataset, consistent with prior publications \cite{jiang2024pocketflow,Ragoza_Masuda_Koes_2022_LIGAN}. The selected complexes correspond to PDB IDs: 1BVR, 1ZYU, 2ATI, 4BNW, 5G3N, 1U0F, 2AH9, 2HW1, 4I91, and 5LVQ. Additionally, we included antibody-antigen pairs sourced from the Structural Antibody Database (SAbDab) for showing the replacement of residues by small molecule fragments \cite{dunbar2014sabdab}.

\subsection{Model overview}
Peptide2Mol is a non-autoregressive diffusion based generative model designed to generate and optimize small molecules within protein pockets, leveraging peptide-binder structural data. As shown in Figure~\ref{fig:overview}a, the model is trained on a curated dataset combining small-molecule conformation ensembles \cite{axelrod2022geom}, protein-ligand/peptide complexes \cite{wang2005pdbbind,zhang2024biolip2}, and protein-peptide models \cite{varadi2022alphafold}. During training, ligand geometries undergo progressive disruption via a diffusion process, while peptide sidechains are partially diffused and binding pocket residues remain fixed. This framework establishes an invertible mapping between the base Gaussian distribution and the ground truth graph. At inference, the model iteratively transfers Gaussian noise into molecular structures at each step until convergence. Non-covalent interactions within 5\text{\AA} are explicitly modeled to capture pocket–ligand contacts (Figure~\ref{fig:overview}b). 

As depicted in Figure~\ref{fig:overview}c, the model embeds ligand and pocket atoms into node and edge features, augmented with timestep embeddings and pocket embedding to distinguish atomic contexts. These representations are processed by six E(3)-equivariant GNN layers that iteratively update atomic features and coordinates through rotation-equivariant convolutions. Finally, the refined embeddings are decoded into molecular graphs. Pocket2Mol \cite{peng2022pocket2mol} can be optionally used to resolve steric clashes to further refine ligand-pocket complementarity.

\subsection{Molecular Featurization}
Ligands and ligand–protein complexes were represented as undirected atomic graphs, denoted as $\mathcal{M} = (\mathcal{V}, \mathcal{E})$. 
Each node $v_i \in \mathcal{V}$ corresponds to an atom and is associated with two attributes: its spatial coordinate $r_i \in \mathbb{R}^3$ and its element-type feature $a_i \in \mathbb{R}^8$, where $a_i$ is implemented as a one-hot encoding over common atom symbols (C, N, O, F, P, S, Cl, Br). 
Each edge $e_{ij} \in \mathcal{E}$ corresponds to an atom pair and is described by a bond feature vector $b_{ij} \in \mathbb{R}^6$, encoding single, double, triple, aromatic, and non-bonded proximity interactions, plus an absorbing state for no interaction. 

\subsection{Diffusion Model Architeture}
 A diffusion model is then employed in the generation process, characterized by two Markov random processes. The forward process incrementally introduces noise to the data according to a predefined noise schedule, while the reverse process leverages neural networks to denoise the data, ultimately reconstructing the original data from the noise. Let the superscript t denote variables at time step t with $\mathcal{M}_0$ representing the 3D molecule or complex drawn from the real distribution. At each step, $\mathcal{M}_t$ is sampled from the conditional distribution $q(\mathcal{M}_t \mid \mathcal{M}_{t-1})$, dependent solely on $\mathcal{M}_{t-1}$:
 
\begin{equation}
    q(\mathcal{M}_t \mid \mathcal{M}_{t-1}, \mathcal{M}_{t-2}, \cdots, \mathcal{M}_0) \coloneqq q(\mathcal{M}_t \mid \mathcal{M}_{t-1})
    \label{markov}
\end{equation}

For atom positions $\mathbf{r}_i$, atom types $a_i$ and bond types $b_{ij}$, which are discrete, categorical distributions are used for their representation. The forward process is defined as:
\begin{align}
    q(\mathbf{r}_i^t \mid \mathbf{r}_i^{t-1}) &\coloneqq \mathcal{N}(\mathbf{r}_i^t \mid \sqrt{1-\beta^t}\mathbf{r}_i^{t-1}, \beta_t\mathbf{I}) \label{r_diffusion} \\
    q(a_i^t \mid a_i^{t-1}) &\coloneqq \mathcal{N}(a_i^t \mid \sqrt{1-\beta^t}a_i^{t-1}, \beta_t\mathbf{I}) \label{a_diffusion} \\
    q(b_{ij}^t \mid b_{ij}^{t-1}) &\coloneqq \mathcal{N}(b_{ij}^t \mid \sqrt{1-\beta^t}b_{ij}^{t-1}, \beta_t\mathbf{I}) \label{b_diffusion}
\end{align}

\noindent with $\beta^t \in [0, 1]$ denotes the predefined noise scaling schedule, $\mathbf{I} \in \mathbb{R}^{3\times3}$  is the identity matrix. For atom position $\mathbf{r}_i$, atom type $a_i$, and bond type $b_{ij}$, scaled standard Gaussian noise is incrementally added. In addition, time embedding and a binary pocket indicator (0/1) were concatenated with the node embeddings, resulting in a unified representation that integrates atomic, temporal, and contextual information.

Leveraging the Markov property, $\mathcal{M}$ can be directly derived from the original sample $\mathcal{M}_0$. By defining $\alpha_t \coloneqq 1 - \beta_t$, and $\bar{\alpha}^t \coloneqq \prod_{s=1}^t \alpha^s$, the sample can be expressed as the following equations:
where $\bar{\alpha}^t = \prod_{s=1}^t (1-\beta^s)$ denotes the fraction of information retained at step $t$. From this formulation, $\bar{\alpha}^t$ represents the proportion of information from the original data retained at step t. We refer to $\bar{\alpha}^t$ as the "information level," which is determined by the noise level $\beta^t$.
\begin{align}
    q(\mathbf{r}_i^t \mid \mathbf{r}_i^0) &\coloneqq \mathcal{N}(\mathbf{r}_i^t \mid \sqrt{\bar{\alpha}^t}\mathbf{r}_i^{0}, (1-\bar{\alpha}^t)\mathbf{I}) \label{rr_diffusion} \\
    q(a_i^t \mid a_i^{0}) &\coloneqq \mathcal{N}(a_i^t \mid \sqrt{\bar{\alpha}^t}a_i^{0}, (1-\bar{\alpha}^t)\mathbf{I}) \label{aa_diffusion} \\
    q(b_{ij}^t \mid b_{ij}^{0}) &\coloneqq \mathcal{N}(b_{ij}^t \mid \sqrt{\bar{\alpha}^t}b_{ij}^{0}, (1-\bar{\alpha}^t)\mathbf{I}) \label{bb_diffusion}
\end{align}

As $t \rightarrow \infty$, the atom positions, types and bond types approximate a standard Gaussian distribution. These resulting prior distributions serve as the initial distributions for the reverse process.

In the reverse process, we invert the Markov chain to reconstruct the original sample from prior distributions, employing E(3)-equivariant neural networks to parameterize the transition probability $p_\theta(\mathcal{M}_{t-1} \mid \mathcal{M}_t)$. Specifically, we model all the three predicted distributions as a Gaussian distribution $\mathcal{N}(\mathbf{X}^{t-1}_i \mid (\mu_\theta(\mathcal{M}_t, t), \Sigma_t)) $, where $\mathbf{X}$ represents variable and $\mu_\theta$ is the neural network. The neural network is trained to recover $\mathcal{M}_{t-1}$ from $\mathcal{M}_t$ by optimizing the predicted distribution $p_\theta(\mathcal{M}_{t-1} \mid \mathcal{M}_t)$ to approximate the true posterior $q(\mathcal{M}_{t-1} \mid \mathcal{M}_t, \mathcal{M}_0)$. During training, the loss function was defined in equations~\eqref{pos_loss}--\eqref{total_loss}.
\begin{align}
    \mathcal{L}_{pos}^{t-1} &= \frac{1}{N}\sum_i \lVert \mathbf{r}_i^{t-1} -\mu_\theta(\mathcal{M}_t, t)_i \rVert ^2 \label{pos_loss}\\
    \mathcal{L}_{atom}^{t-1} &= \frac{1}{N}\sum_i \lVert a_i^{t-1} -\mu_\theta(\mathcal{M}_t, t)_i \rVert ^2 \label{atom_loss} \\
    \mathcal{L}_{bond}^{t-1} &= \frac{1}{N}\sum_i \lVert b_{ij}^{t-1} -\mu_\theta(\mathcal{M}_t, t)_i \rVert ^2 \label{bond_loss} \\
    \mathcal{L}^{t-1} &= \mathcal{L}_{pos}^{t-1} + \lambda_a \mathcal{L}_{atom}^{t-1} + \lambda_b \mathcal{L}_{bond}^{t-1} \label{total_loss}
\end{align}
where $\lambda_a$ and $\lambda_b$ were set both 30 for atom and bond. A timestep t was randomly sampled during training and the neural network was applied to recover the unbiased molecule, where its parameters was optimized by minimizing the loss $\mathcal{L}^{t-1}$. In inference process, we sample atom position, type and symmetric bond type in Gaussian distribution and repeatedly sample from $t = T, T-1, \cdots, 1$ to denoise the molecule.

\subsection{Molecule Generation}
Inference process is used for the generation of small molecules. During inference, the model takes the receptor pocket originally defined by the peptide–protein interface but does not include the peptide backbone as a structural scaffold. Instead, the model initializes from Gaussian noise and iteratively denoises the atomic positions, atom types, and bond connectivity to generate a small molecule directly within the peptide’s binding pocket. In this way, the generated molecules adopt drug-like geometries while preserving the essential interaction pattern of the original peptide because of the diverse training data. This design allows Peptide2Mol to effectively translate peptide binding interfaces into small molecule mimetics rather than reconstructing peptide structures. Peptide2Mol can also make molecule optimization and peptidomimetic design when pointing fixed atoms during diffusion process.

After generation, a pocket-aware refinement stage can be applied using the Pocket2Mol optimization module \cite{peng2022pocket2mol}. This step performs local atom and bond adjustment to remove steric clashes, correct unreasonable geometries, and improve shape complementarity between the ligand and pocket. Such refinement is necessary because diffusion sampling may yield high-energy or overlapping conformations that are not physically realizable. The Pocket2Mol-based relaxation ensures that the final small molecules correspond to chemically valid, energetically stable binding poses consistent with the protein pocket environment.

\subsection{Fragmentation of Small Molecules}
To identify which fragments were most frequently used to replace residue side chains, we filtered the SabDab dataset for complementarity-determining region (CDR) domains in complex with antigens. Antigens were defined as residues within 5\text{\AA} of the CDR domain, and only complexes where the number of antigen residues exceeded the number of CDR residues were retained. These CDRs were then converted into small molecules using Peptide2Mol, and the resulting molecules were fragmented based on their rotatable bonds. A fragment was defined as replacing an amino acid if it was located within 4~\text{\AA} of any heavy atom of the residue. To rank the likelihood of fragment–residue replacements, we computed the Pointwise Mutual Information (PMI) as follows:

\begin{align}
\textbf{PMI}(\text{res}, \text{frag}) = \log \left( \frac{P(\text{res}, \text{frag})}{P(\text{res}) \cdot P(\text{frag})} \right)
\end{align}

\section{Results}
\subsection{Benchmarking Molecular Properties of Peptide2Mol}
We first assessed the general properties of molecules generated by Peptide2Mol, focusing on evaluating their chemical validity and plausibility. To this end, we selected and computed the following evaluation metrics, which have been widely adopted in previous studies to characterize the properties of sampled candidates \cite{peng2022pocket2mol,peng2023moldiff,zhou2024reprogramming}: (1) QED  (Quantitative Estimation of Drug-likeness) \cite{bickerton2012quantifying}, quantifies the likelihood of a molecule being a viable drug candidate based on its physicochemical properties and conformity to drug-like characteristics; (2) SA (Synthetic Accessibility) \cite{ertl2009estimation}, measuring the ease of molecular synthesis, with higher scores indicating greater synthetic feasibility; (3) LogP (Octanol–Water Partition Coefficient), a metric of molecular hydrophobicity derived from the distribution between octanol and aqueous phases; and (4) PBrate (PoseBusters passing rate)  \cite{buttenschoen2024posebusters} integrates 19 criteria to comprehensively assess docking quality, including molecular structural integrity and conformational validity, which provides a rigorous and comprehensive measure of docking plausibility and makes it a reliable benchmark for assessing generative models.

\begin{table}[htb]
\caption{The comparison of properties of the generated molecules in the test set.}
\vspace{5pt}
\label{tab:performance}
\centering
\begin{tabular}{
|>{\centering\arraybackslash}p{4cm}|
 >{\centering\arraybackslash}p{1.5cm}|
 >{\centering\arraybackslash}p{1.5cm}|
 >{\centering\arraybackslash}p{1.5cm}|
 >{\centering\arraybackslash}p{2.2cm}|}
\hline
\textbf{Method} & \textbf{QED($\uparrow$)} & \textbf{SA($\uparrow$)} & \textbf{LogP} & \textbf{PBrate(\%. $\uparrow$)} \\
\hline
LiGAN \cite{Ragoza_Masuda_Koes_2022_LIGAN} & 0.428 & 0.546 & 1.224 & 39.50 \\
Pocket2Mol \cite{peng2022pocket2mol} & \textbf{0.587} & 0.758 & 1.063 & 71.60 \\
TargetDiff \cite{guan2023targetdiff} & 0.430 & 0.550 & 1.249 & 36.90 \\
PocketFlow \cite{jiang2024pocketflow} & 0.497 & \textbf{0.769} & 3.521 & 46.00 \\
Peptide2Mol & 0.501 & 0.612 & 0.638 & 45.30 \\
Peptide2Mol-Fixed & 0.509 & 0.637 & 0.729 & \textbf{83.80} \\
\hline
\end{tabular}
\end{table}

\textbf{Table}~\ref{tab:performance} summarizes the generative performance of Peptide2Mol compared with representative molecular generation methods on the same benchmark used in LiGAN \cite{Ragoza_Masuda_Koes_2022_LIGAN} and PocketFlow \cite{jiang2024pocketflow}. In terms of QED, our model (0.501) already surpasses LiGAN (0.428) and TargetDiff (0.430), and achieves a comparable level to PocketFlow (0.497), situating it within a competitive range. For SA, while peptide-like molecules naturally score lower compared to small-molecule–oriented methods such as Pocket2Mol and PocketFlow, Peptide2Mol maintains parity with LiGAN and TargetDiff, highlighting its balance between peptide-specific features and overall synthetic feasibility. Regarding hydrophobicity, the lower LogP values generated by Peptide2Mol reflect the intrinsic physicochemical properties of peptides, making the results consistent with the intended design space. 

\begin{figure}[htb]
\vspace{-14pt}
    \centering
    \includegraphics[width=0.9\linewidth]{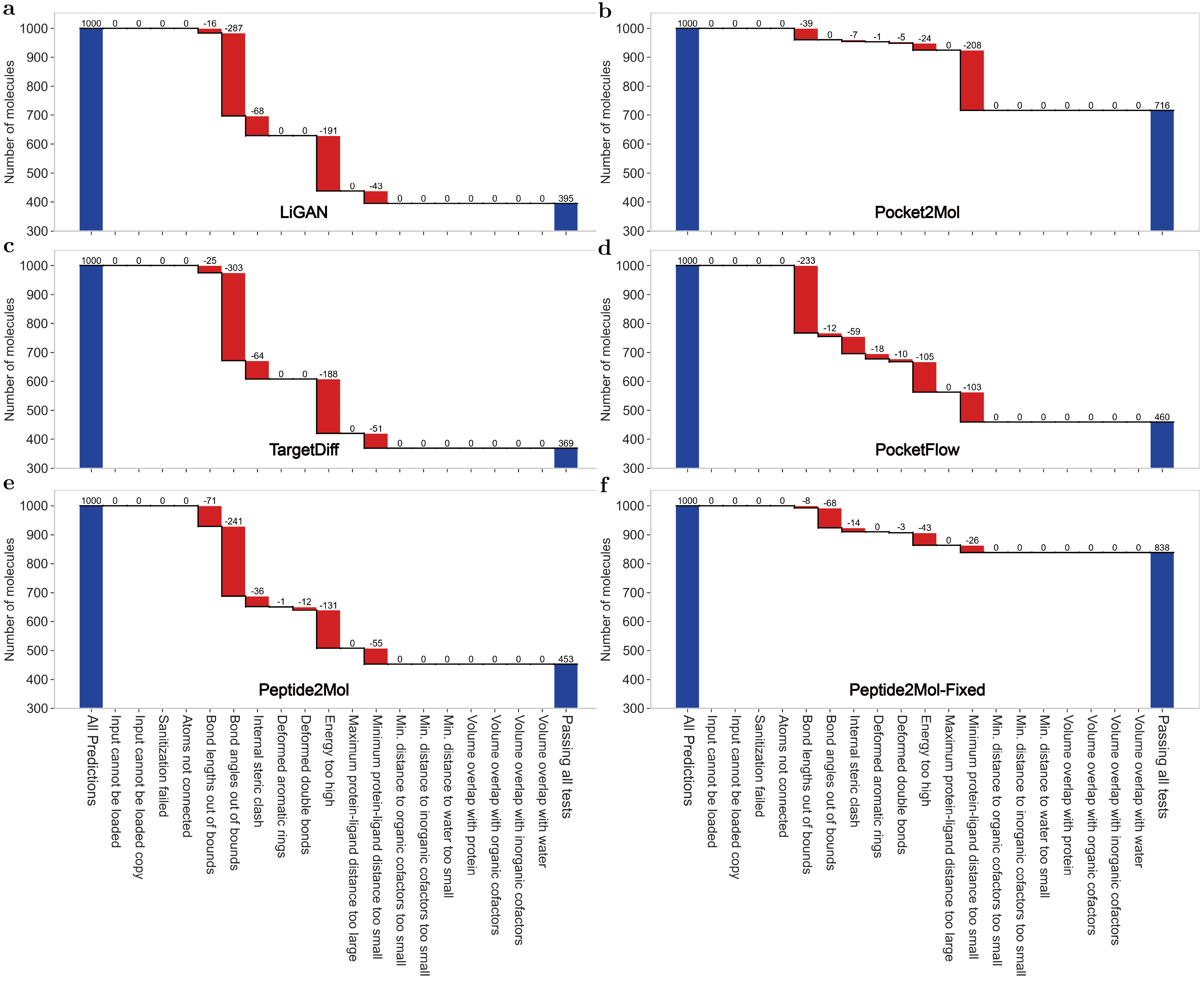}
    \caption{\textbf{Waterfall diagram illustrating the stepwise evaluation of AI-generated molecules against the PoseBusters criteria.} Each method was designed to generate 100 molecules per target across the testset targets. Panels show results for LiGAN (\textbf{a}), Pocket2Mol (\textbf{b}), TargetDiff (\textbf{c}), PocketFlow (\textbf{d}), Peptide2Mol (\textbf{e}), and Peptide2Mol-Fixed (\textbf{f}).}
    \label{fig:posebusters}
\end{figure}

Although Peptide2Mol does not achieve the highest scores in QED or SA, its performance is comparable to established approaches trained exclusively on small-molecule datasets. Importantly, when a partially masked autoregressive refinement step is applied (Peptide2Mol-Fixed), the overall chemical validity is further improved, yielding the highest PBrate (83.80\%). This indicates that molecules generated by Peptide2Mol, although not optimized exclusively for drug-likeness metrics, achieve competitive quality and can be effectively refined to ensure robust structural integrity and docking plausibility.

To further dissect the structural quality of the generated molecules, we visualize the individual contributions to the PoseBusters score using a waterfall plot (Figure \ref{fig:posebusters}). This representation highlights which specific geometric and conformational criteria most strongly influence the PBrate for each method. For instance, Pocket2Mol demonstrates strong performance in satisfying bond length distribution constraints and in generating molecules with favorable internal energies. In contrast with Pocket2Mol, Peptide2Mol achieves superior control over intermolecular distance constraints with the target, thereby effectively reducing steric clashes. Leveraging these complementary strengths, we employed Pocket2Mol for partially-refinement of our generated molecules, which yielded the most favorable overall evaluation outcomes.

\subsection{Bond Length Distribution Analysis}
\begin{figure}[H]
    \centering
    \includegraphics[width=0.9\linewidth]{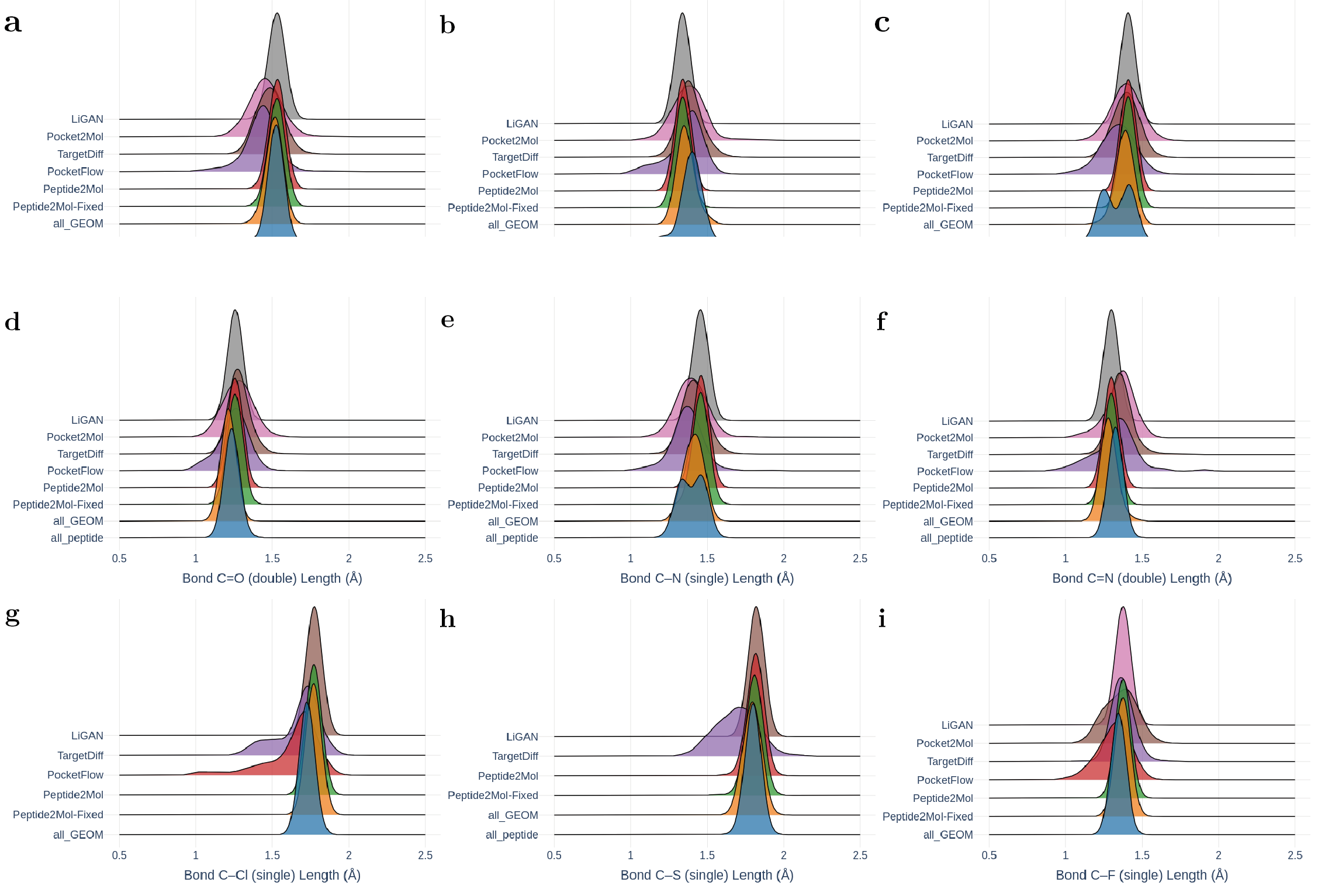}
    \caption{\textbf{Geometric and property-based evaluation of generated molecules.}
    \textbf{(a–i)} Bond length distributions of molecules generated by different AI-based methods compared with those in the training set. Nine representative bond types are analyzed: C–C (\textbf{a}), C=C (\textbf{b}), C–O (\textbf{c}), C=O (\textbf{d}), C–N (\textbf{e}), C=N (\textbf{f}), C–Cl (\textbf{g}), C–S (\textbf{h}), and C–F (\textbf{i}). }
    \label{fig:bond_and_metrics}
\end{figure}

In addition to the benchmark comparison, we further examined the bond length distributions of generated molecules. As shown in Figure~\ref{fig:bond_and_metrics}a-i, Nine kinds of chemical bonds are analyzed, including C-C, C=C, C-O, C=O, C-N, C=N, C-Cl, C-S, and C-F. 

Notably, the results from Peptide2Mol closely match the overall distribution of the training dataset, while also notably capturing the characteristic bond length patterns specific to peptides. This highlights the model’s ability to generate peptide-like molecules that are both chemically realistic and structurally consistent with experimental observations. 

\subsection{Residue replacement analysis}
\begin{figure}[htb]
   \centering
   \includegraphics[width=0.8\linewidth]{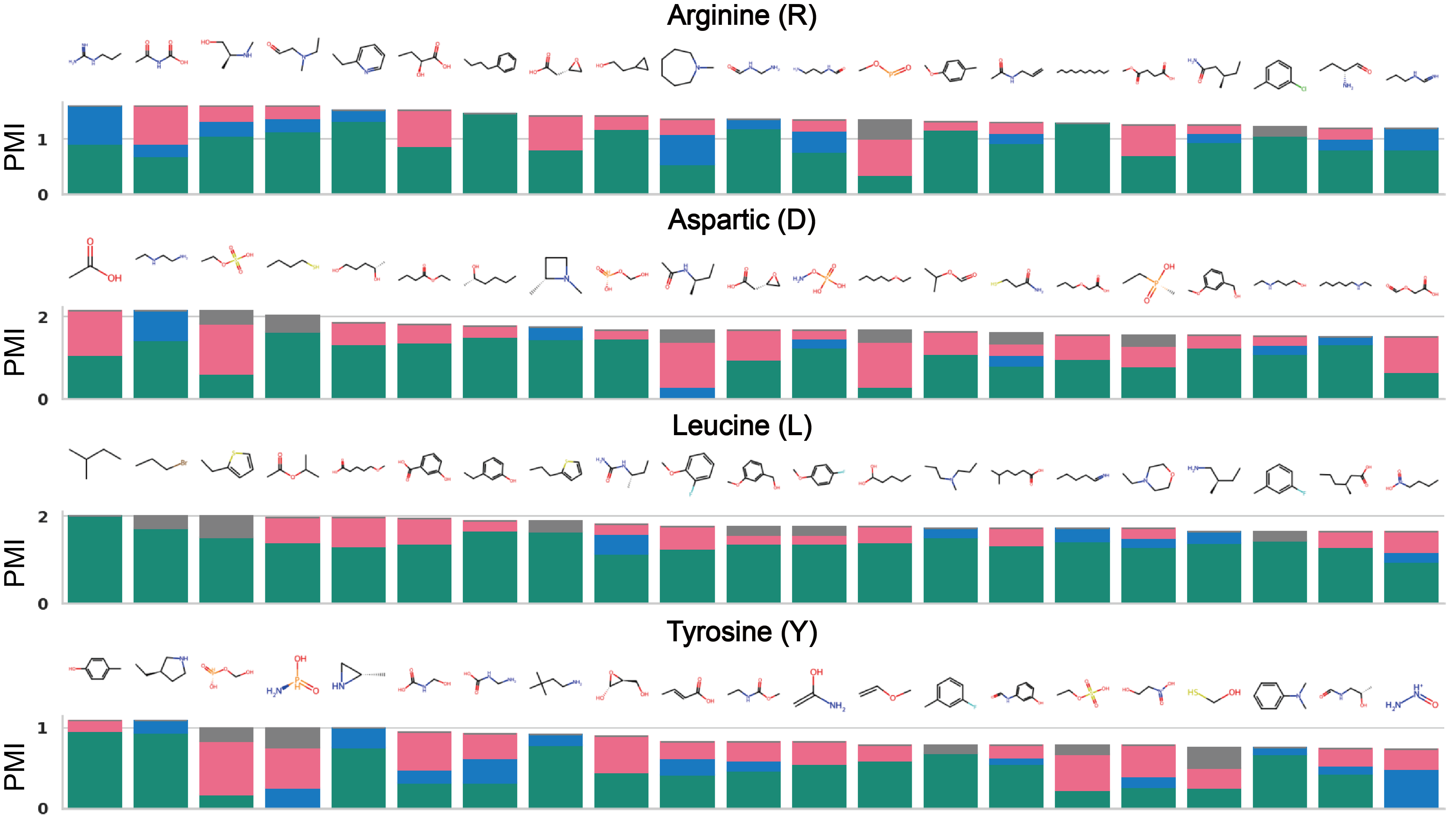}
   \caption{The histogram to show the top replacement fragment from small molecules with 4 representative residues (ARG, ASP, LEU and TYR), the color reflects the composition proportion of elements (green: Carbon, Blue: Nitrogen, Red: Oxygen, Gray, others)}
   \label{fig:residue_mimicry}
   \vspace{-25pt}
\end{figure}

To investigate the residue-level mimicry capability of Peptide2Mol, we applied the model to an antibody–antigen dataset to generate small-molecule fragments substituting native antibody side chains. Four representative residues—tyrosine (Y), aspartic acid (D), arginine (R), and leucine (L)—were analyzed (Figure~\ref{fig:residue_mimicry}). Fragments were ranked by PMI with the corresponding residue, reflecting association strength. High-PMI fragments generally preserve key chemical features: tyrosine substitutes retain aromatic or hydroxyl groups; aspartic acid replacements are enriched in polar oxygen-containing groups; arginine mimics maintain nitrogen-rich motifs; and leucine substitutes comprise carbon-rich hydrophobic chains. These results indicate that Peptide2Mol generates chemically plausible, residue-specific side-chain mimics while allowing structural diversity.

\section{Discussion}
\begin{figure}[htb]
\centering
\includegraphics[width=0.8\linewidth]{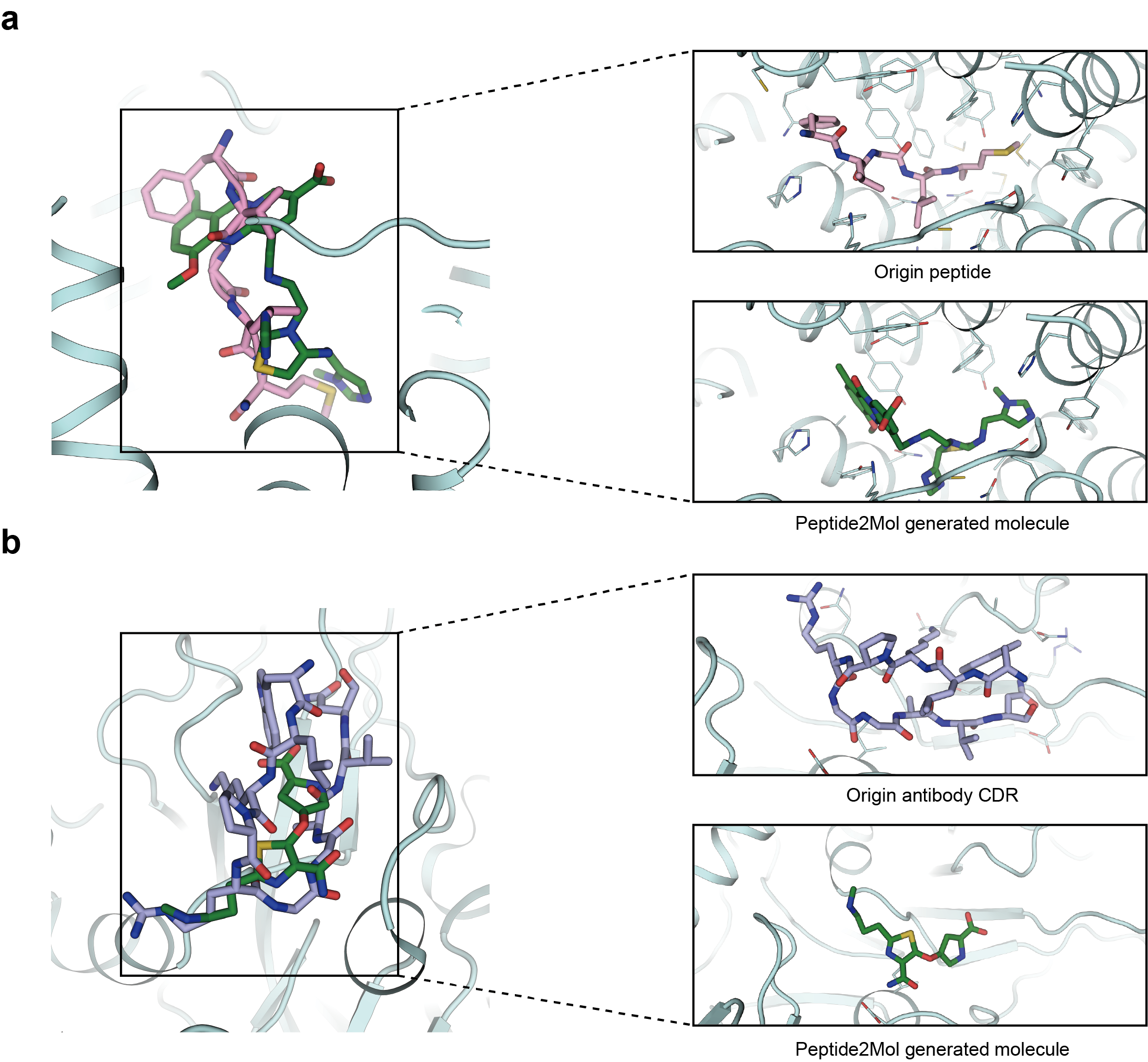}
\caption{Representative examples showing that Peptide2Mol can transform (a) a peptide binder (PDB ID: 7WXO) and (b) an antibody CDR (PDB ID: 3NGB) into corresponding small molecules that mimic their binding interfaces.}
\label{fig:f4}
\vspace{-15pt}
\end{figure}

In this work, we introduced Peptide2Mol, a structure-based generative framework designed to bridge the gap between peptides and small molecules in drug discovery. Unlike previous generative models that primarily focus on small-molecule–protein complexes, Peptide2Mol integrates structural information from both protein–ligand and protein–protein (or protein–peptide) interactions. This enables the model to translate peptide or antibody CDR binders into small molecules that mimic their native binding modes (Fig. \ref{fig:f4}). This design enables the generation of peptide-mimicking small molecules that preserve the functional essence of native residues while retaining drug-like chemical properties or generate peptidomimetics from original peptide.

One strength of Peptide2Mol is its principled use of diverse structural datasets in training. Existing models often inherit biases from protein–ligand complexes \cite{jiang2024pocketflow,peng2022pocket2mol,peng2023moldiff,zhou2024reprogramming}. By systematically incorporating both experimentally determined and computationally predicted peptide and protein interaction data, Peptide2Mol broadens the generative chemical space. This approach improves the diversity of generated molecules, while still yielding competitive performance in standard benchmarks. Importantly, refinement with a partially masked autoregressive step significantly improved structural plausibility, achieving the highest PoseBusters passing rate, thereby demonstrating the potential of combining complementary modeling strategies.

Despite these advances, several limitations remain. The generated molecules tend to inherit physicochemical features closer to peptides than to conventional small molecules, which may explain their modest performance on QED and SA relative to models optimized exclusively for drug-likeness. While this aligns with the goal of peptide mimicry, practical applications will require balancing peptide-like fidelity with pharmacokinetic constraints \cite{datta2019mechanisms}. Moreover, although we demonstrated residue-level replacement analysis, the current model does not yet provide a quantitative metric for peptide–small molecule similarity.

Looking forward, we envision several directions for extending this work. One is to couple Peptide2Mol with physics-based simulation pipelines to assess stability and binding mechanisms beyond docking scores \cite{van2023physics}. Moreover, systematic benchmarking across a broader range of “undruggable” protein–protein interaction targets will be critical to establish the generality of this approach and to uncover patterns of residue substitution that may inform rational drug design \cite{sun2025computer}.

In conclusion, Peptide2Mol represents a step toward unifying peptide- and small-molecule–based design strategies. By capturing the structural logic of peptide binders while ensuring drug-like feasibility, our framework highlights a new frontier for generative drug discovery. Just as the development of protein language models expanded the interpretability of sequence variation, the integration of peptide-derived binding information into molecular generation holds promise to unlock new chemical modalities and accelerate the translation of peptide insights into therapeutically viable small molecules.

\section{Code Availability}
The source code, pretrained models, and a minimal test dataset for Peptide2Mol are publicly available at \url{https://github.com/BLUE-Flowing/Peptide2Mol/}.

%
%
%
%

\end{document}